\documentclass{article}

\usepackage{microtype}
\usepackage{graphicx}
\usepackage{subcaption}
\usepackage{booktabs} 

\usepackage{hyperref}

\usepackage[accepted]{icml2026}

\usepackage{mathtools}
\usepackage{amsthm}

\usepackage[capitalize,noabbrev]{cleveref}

\usepackage{amsmath,amssymb}
\usepackage{graphicx}
\usepackage{booktabs}
\usepackage{array}
\usepackage{multirow}
\usepackage{microtype}
\usepackage{tikz}
\usepackage{hyperref}
\hypersetup{
    colorlinks=true,
    linkcolor=blue,
    citecolor=blue,
    urlcolor=blue
}
\usetikzlibrary{shapes.geometric,arrows.meta,positioning}

\icmltitlerunning{Context-Aware Hierarchical Bayesian Analysis of IVF Environmental Conditions}

\begin{document}

\twocolumn[
\icmltitle{Context-Aware Hierarchical Bayesian Modeling of\\ IVF Laboratory Environmental Conditions}

 \begin{icmlauthorlist}
    \icmlauthor{Zahra Asghari Varzaneh}{mau}
    \icmlauthor{Reza Khoshkangini}{mau}
\icmlauthor{Pia Saldeen}{IVF}
    \icmlauthor{Lars Johansson}{AB} 
    \icmlauthor{Thomas Ebner}{kuk}
    
  \end{icmlauthorlist}

  \icmlaffiliation{mau}{Department of Computer Science and Media Technology,
    Malmö University, Malmö, Sweden}
    \icmlaffiliation{IVF}
  {Nordic IVF, Sweden}
  \icmlaffiliation{AB}
  {NewLifeAid-Global AB, Sweden}
  \icmlaffiliation{kuk}{Kepler Universitätsklinikum, Linz, Austria}

  \icmlcorrespondingauthor{Zahra Asghari Varzaneh}{zahra.asghari-varzaneh@mau.se}

\icmlkeywords{IVF, context-aware features, hierarchical Bayes,
              Beta regression, partial pooling, SHAP, health informatics}
\vskip 0.3in
]

\printAffiliationsAndNotice{}

\begin{abstract}
IVF pregnancy rates are routinely modeled using patient-level variables, while high-resolution laboratory environmental data remain underutilized. We show that this is a missed opportunity. Rather than relying on raw sensor averages, we engineer 55 context-aware temporal features including rolling thermal stability, simultaneous temperature-humidity adherence, peak stress duration, and post-stress recovery speed that capture the dynamics of incubator microenvironments. On 61 weeks of data from an Asian IVF clinic, these features reduce cross-validated prediction error to 1.27\%, compared to 3–5\% for raw averages. We then train a hierarchical Bayesian Beta regression model that shares environmental effects across an Asian and a Northern European clinic via partial pooling, while preserving site-specific baselines. On held-out data from the Northern European clinic, the model achieves R² = 0.86 and a 64\% error reduction for the 35–39 age group over a naive baseline,demonstrating that structured environmental monitoring contains clinically meaningful, transferable signal.
\end{abstract}
\section{Introduction}
The success of IVF depends on several interacting factors, including the
patient's age, ovarian reserve, sperm quality, lifestyle, the stimulation
protocol, embryo quality, etc~\citep{younglai2005}. In addition, the physical environmental
conditions of the laboratory, like temperature, humidity, and air quality, play
an important role. During the culture period, embryos are kept inside
incubators that tightly control their internal environment~\citep{mantikou2013,varzaneh2025}. However, these
incubators do not operate in complete isolation. Each time an incubator door
is opened, room air enters, and the time needed for the incubator
to return to its optimal condition depends directly on how stable the
surrounding laboratory environment is.
Despite this, environmental monitoring in IVF laboratories is often treated primarily as a compliance exercise: sensors trigger alerts when fixed thresholds are breached, yet the rich, high-resolution sensor data are rarely analyzed statistically. This represents a missed opportunity. Prior
work has focused on patient-level predictors~\citep{kupka2016}, suggesting the influence of environmental and procedural factors. Complementary controlled experiments, such as ~\citep{swain2012}, demonstrate that even minor changes in culture conditions, such as oxygen levels, temperature stability, and media pH, can directly affect embryo development and clinical outcomes. However, these studies typically examine single variables under idealized conditions and do not capture the complex, time-varying fluctuations present in routine clinical practice. As a result, ignoring high-resolution environmental data may miss the cumulative effects of real-world stressors that help explain differences between laboratories.\\
We address this gap with two contributions: \\
\textbf{First}, we introduce \emph{context-aware features} computed from sensor data that
encode temporal patterns invisible to raw statistics: rolling thermal
stability, the fraction of time temperature and humidity are
simultaneously within their biological ideal ranges, the longest
uninterrupted stress episode, post-stress recovery speed, and first week
transfer-window summaries. \\
\textbf{Second}, we develop a \textit{hierarchical Bayesian Beta regression} model to analyze IVF outcomes across two clinics (Asian and Northern European). The model uses partial pooling to share information about environmental effects between the two sites, while allowing each clinic to maintain its own baseline level. In this way, data from the Asian clinic helps the model learn general and stable relationships between environmental conditions and IVF outcomes, which in turn improves prediction performance on the Northern European dataset without assuming that the two clinics are identical.\\
This study also address three key questions:\\
\textbf{(i)} Do context-aware environmental features outperform simple aggregated sensor measures?\\
\textbf{(ii)} Which features are most important for prediction, and are these effects consistent across the two clinics?\\
\textbf{(iii)} Does sharing information between clinics through partial pooling improve performance compared to a single-site?

\section{Data}

Environmental sensor data were collected at two IVF laboratories at 10-minute intervals
during 2024--2025. We refer to the Asian clinic as the \textbf{source} clinic because it
provides substantially more labelled data (61~weeks) and serves as the basis for
feature selection and partial-pooling priors; the Northern European clinic is the
\textbf{target}, representing the site we aim to generalise to. Both stations record temperature (°C),
relative humidity (\%), CO$_2$ (ppm), and TVOC (ppb). Pregnancy rates were provided for
patients under~35, aged 35--39, and aged 40 and above. Table~\ref{tab:env} summarises the training period.
The Asian record spans January~2024--October~2025, interrupted by a
196-day collection gap caused by sensor data unavailability at the clinic; this splits
it into two continuous segments of 19~and 42~weeks (61~total). The Northern European clinic provides 14~months
(November~2024--December~2025). The large standard deviations in the Northern European data mostly come from months with only one to three patients in an age group, where the reported rates are driven more by small numbers than by environmental conditions.
\vspace{-0.9em}
\paragraph{Data structure.}
Environmental variables are recorded every 10 minutes. Pregnancy rates were reported at
different temporal granularities by the two clinics: weekly for Asia and monthly for Northern
Europe, reflecting the format in which outcome data were made available to us by each
laboratory. Environmental features are therefore aggregated to match each clinic's reporting
interval (weekly for Asia, monthly for Northern Europe), so that features and outcomes are
always aligned at the same resolution. The 196-day gap in the Asian data is handled by
treating the two segments as independent time series during cross-validation, with fold
boundaries never placed across the gap to avoid information leakage; the autoregressive term
$\gamma\,\mathrm{logit}(y_{t-1,c})$ is reset at the start of each segment.
A further challenge is that outcomes at time~$T$ reflect conditions from roughly
two to six weeks earlier. Since exact embryo-level timestamps are unavailable,
we approximate this delay by including lagged environmental summaries from the
previous period ($T-1$), providing a practical alignment between past conditions
and observed outcomes.

\begin{table}[b]
\centering
\caption{Environmental conditions and pregnancy rates: mean$\pm$std,
training period.}
\label{tab:env}
\small
\begin{tabular}{lrrl}
\toprule
Variable & Asia & Northern Europe & Unit \\
\midrule
\multicolumn{4}{l}{\textit{Environmental}} \\
Temperature  & $23.4\pm0.6$ & $24.0\pm0.6$ & °C \\
Humidity     & $37.2\pm9.8$ & $42.8\pm8.3$ & \% \\
CO$_2$       & $168\pm22$   & $189\pm18$   & ppm \\
TVOC         & $148\pm420$  & $312\pm380$  & ppb \\
\midrule
\multicolumn{4}{l}{\textit{Pregnancy rates by age group}} \\
PR ($<$35 yrs)   & $44.1\pm3.1$  & $38.6\pm18.4$ & \% \\
PR (35--39 yrs)  & $26.6\pm2.8$  & $34.7\pm12.1$ & \% \\
PR (40+ yrs)     & $13.9\pm1.8$  & $16.2\pm13.7$ & \% \\
\bottomrule
\end{tabular}
\end{table}

\section{Method}
\begin{figure}[t]
\centering
\includegraphics[width=\columnwidth, trim=0cm 0cm 0cm 0cm, clip]{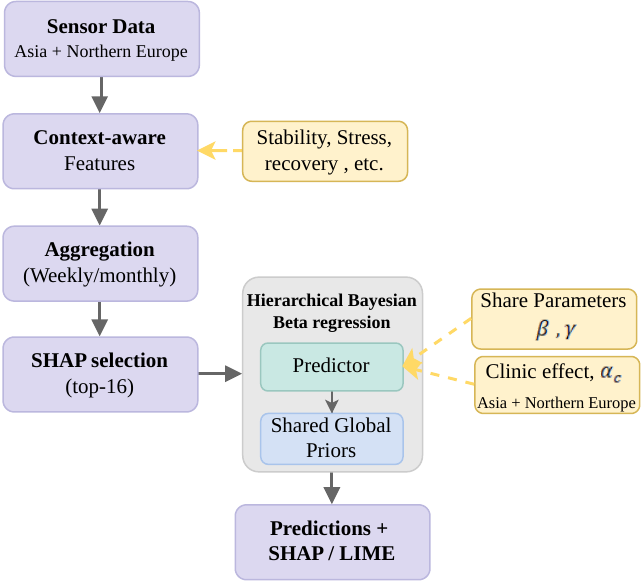}
\caption{From sensor data to predictions. Context-aware features are extracted and aggregated over time, then fed into a hierarchical Bayesian model that shares information across clinics while preserving site-specific differences}
\label{fig:flow}
\end{figure}

The pipeline has four steps (see Figure~\ref{fig:flow}): (1)~context-aware
feature engineering from 10-minute readings; (2)~aggregation to weekly
resolution for Asia and monthly for Northern Europe; (3)~SHAP-based selection of the
top-16 features per age group on Asian data to keep sampling tractable;
(4)~joint modelling of both clinics using a hierarchical Bayesian Beta regression.
\paragraph{Step 1: Context-aware features.}
For temperature $T_t$ with optimum $T^*=22.5$\,°C, the stress index is
$s^T_t = \max(0,\,|T_t-T^*|-1.5)/3$; an analogous expression governs
humidity ($H^*=42.5\%$, deadband $\pm2.5\%$). The composite comfort index
$c_t=(1-0.5\,s^T_t-0.5\,s^H_t)^+$ summarises environmental quality. Beyond
these pointwise quantities, we compute: the rolling 1-hour temperature
standard deviation $\sigma^T_t$ (and its period mean and stable fraction
below 0.5\,°C) as a measure of thermal stability; the ideal-zone adherence
$z_t=\mathbf{1}[T_t\!\in\![21,24]^{\circ}\mathrm{C}]\cdot
\mathbf{1}[H_t\!\in\![40,45]\%]$ capturing simultaneous optimality; the
longest consecutive stress episode; the recovery score
$r=\mathbb{E}[c_{t+1}\mid c_t<0.5]$; Short-term lags (1--3~hours) capture local dynamics, while summaries from the preceding period ($T-1$) provide coarse temporal alignment with outcomes. The final feature set includes 55 context-aware variables per period.

\textbf{Step~2: Temporal aggregation.}
Asia segments are aggregated to weekly resolution independently; the Northern European
data to a monthly resolution. This yields 55 features per period.

\textbf{Step~3: SHAP feature selection.}
Given the small number of Northern European training months, using all features would lead
to an unstable Bayesian model. We therefore train an XGBoost model ~\citep{chen2016xgboost} on the Asian
data and select the top-16 features based on mean absolute SHAP importance for
each age group ~\citep{lundberg2017unified}. This reduces dimensionality while retaining the most informative
predictors.
\paragraph{Step~4: Hierarchical Bayesian Beta regression }~\citep{bhuwalka2023}.\label{sec:model}
Let $y_{t,c}\in(0,1)$ denote the pregnancy rate for clinic $c$ at time $t$. Since the outcome is a proportion bounded between 0 and 1, we model it using a Beta distribution:
\begin{equation}
y_{t,c}\sim\mathrm{Beta}(\mu_{t,c}\phi,\,(1-\mu_{t,c})\phi).
\label{eq:beta}
\end{equation}

The expected value $\mu_{t,c}$ is linked to the predictors through:
\begin{equation}
  \eta_{t,c} = \alpha_c
             + \mathbf{x}_t^{\!\top}\boldsymbol{\beta}
             + \gamma\,\mathrm{logit}(y_{t-1,c}),
  \quad
  \mu_{t,c} = \sigma(\eta_{t,c}).
\label{eq:linear}
\end{equation}

As shown in Eq.~\eqref{eq:linear}, $\alpha_c$ is a clinic-specific intercept that captures baseline differences in pregnancy rates between clinics. The coefficient vector $\boldsymbol{\beta}$ is shared across clinics and represents the common effect of environmental features. The autoregressive term $\gamma\,\mathrm{logit}(y_{t-1,c})$ accounts for temporal dependence, reflecting that current outcomes are influenced by previous ones.\\
To enable partial pooling across clinics, we place a hierarchical prior on the intercepts:
\begin{equation}
\alpha_c\sim\mathcal{N}(\mu_G,\sigma_G).
\label{eq:hier}
\end{equation}
where $\mu_G$ is informed by approximate age-specific clinical pregnancy rates
from the general IVF literature (Appendix~A), and $\sigma_G$ controls how much
clinics may deviate from this shared baseline. $\sigma_G$ is inferred from data,
yielding automatic partial pooling without a fixed pooling decision.\\
We use weakly informative priors for the remaining parameters in Eq.~\eqref{eq:priors}. Posterior inference uses NUTS~\citep{hoffman2014} in PyMC~\citep{salvatier2016}
(2~chains, 1500~warm-up\,+\,1500~draws, target acceptance~0.95).
\begin{equation}
\begin{split}
\boldsymbol{\beta}\sim \mathcal{N}(\mathbf{0}, 0.09I), \\
\gamma\sim \mathcal{N}(0.3, 0.09), \\
\phi\sim \mathrm{Gamma}(3, 0.1)
\end{split}
\label{eq:priors}
\end{equation}
\paragraph{Feature selection for the Bayesian model.}
With 11 training months and 55 features, directly fitting all predictors
yields a non-identified posterior. We select the top-16 features per age group
by mean absolute SHAP importance computed on the Asian data (via XGBoost).
\section{Results}

\paragraph{Setup.}
Northern European months~1--11 are used for training; months~12--14
(October--December~2025) are held out. The decision to use Asia as source and Northern
Europe as target rather than a location-independent split is motivated by the
asymmetry in available labelled data: the Asian clinic provides 61~weeks sufficient for
cross-validation, while the Northern European clinic has only 14~months in total. A
location-independent split would mix sites within folds, preventing evaluation of
cross-site transfer, which is the central claim of this work. We compare the hierarchical model
(HM) with a single-site XGBoost baseline (XGB) and a naive predictor
(train-mean). All results use the held-out test set only. The models are evaluated using CV-MAE, MAE, and $R^2$.

\begin{figure}[t]
\centering
\includegraphics[width=\columnwidth]{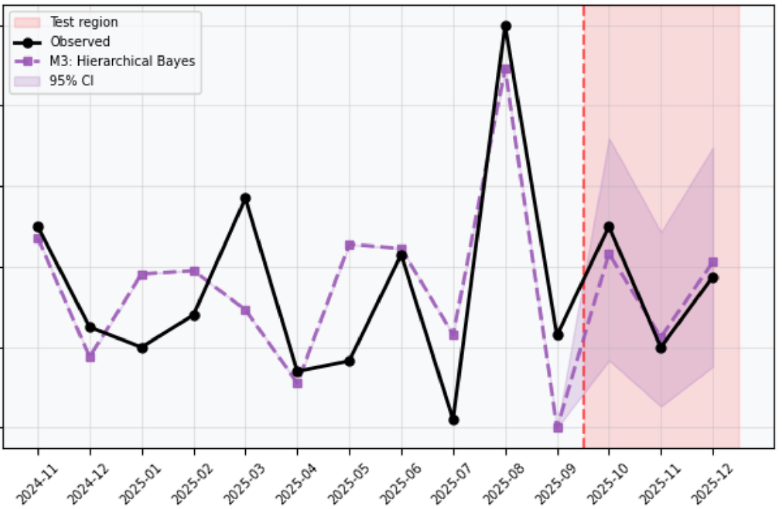}
\caption{PR 35--39: observed (black) vs.\ HM posterior mean (dashed).
Shading = 95\% credible interval. Pink = test period.}
\label{fig:pred}
\end{figure}

\paragraph{Asian cross-validation.}
Table~\ref{tab:iran} reports 5-fold time-series cross-validation on the
Asian data. Context-aware features achieve CV-MAE of 0.85--1.30\%, compared
with 1.57--1.94\% when using only the four raw monthly statistics (mean
temperature, humidity, CO$_2$, TVOC). This consistent improvement across all
age groups indicates that temporal patterns provide useful predictive
information beyond simple averages.

\begin{table}[t]
\centering
\caption{Asian clinic --- 5-fold time-series CV (61~weeks).
Context-aware vs.\ raw-feature baseline (in parentheses).}
\label{tab:iran}
\small
\begin{tabular}{lccc}
\toprule
Age group & Model & CV-MAE & (Raw) \\
\midrule
PR $<$35 yrs  & RF & 1.25\% & (1.94\%) \\
PR 35--39 yrs & RF & 0.85\% & (1.57\%) \\
PR 40+ yrs    & RF & 1.30\% & (1.83\%) \\
\bottomrule
\end{tabular}
\end{table}

\paragraph{Northern European test results.}
Table~\ref{tab:sweden} and Figure~\ref{fig:pred} report test-set metrics. 
\begin{table}[b]
\centering
\caption{Northern European clinic --- test-set results (3 held-out months).
$\Delta$N = improvement over naive. \textbf{Bold} = best MAE per group.}
\label{tab:sweden}
\small
\setlength{\tabcolsep}{3pt}
\begin{tabular}{llccc}
\toprule
Age group & Model & MAE & $R^2$ & $\Delta$N \\
\midrule
\multirow{3}{*}{\shortstack[l]{PR $<$35\\(noise-limited)}}
  & Naive & 8.76\%  & ---    & ---      \\
  & XGB    & 17.05\% & $-5.09$ & $-95\%$ \\
  & \textbf{HM} & \textbf{16.55\%} & $-8.81$ & $-89\%$ \\
\midrule
\multirow{3}{*}{\shortstack[l]{PR 35--39\\\textbf{(primary)}}}
  & Naive & 11.80\% & ---     & ---   \\
  & XGB    & 11.80\% & $-0.28$ & $0\%$ \\
  & \textbf{HM} & \textbf{4.30\%} & $\mathbf{+0.86}$ & $\mathbf{+64\%}$ \\
\midrule
\multirow{3}{*}{\shortstack[l]{PR 40+\\(few patients)}}
  & Naive & 20.14\% & ---     & ---    \\
  & XGB    & 19.85\% & $-0.33$ & $+1\%$ \\
  & \textbf{HM} & \textbf{16.71\%} & $\mathbf{+0.08}$ & $\mathbf{+17\%}$ \\
\bottomrule
\end{tabular}
\end{table}
HM achieves MAE~$=$~4.30\% and $R^2=0.86$ for the 35--39 group, a
64\% improvement over naive. This group is the most reliable target
because it has the largest per-month patient count at both clinics,
making the observed rate a stable estimator. We caution that with only
3 test months the reported point estimates carry substantial uncertainty;
these results should be interpreted as a promising preliminary signal
rather than a definitive claim.
The strongly negative $R^2$ values for XGBoost ($-5.09$, $-0.28$, $-0.33$) indicate that a point-estimate model without regularisation overfits the 11 training months and collapses on the test set as a direct consequence of the limited sample size. This pathological behaviour is precisely what motivates the hierarchical Bayesian approach: the informative prior and partial pooling act as regularisation, preventing overfitting in the low-data regime. For the 40+ group HM
recovers $R^2=+0.08$ and a 17\% improvement; the hierarchical prior
stabilises predictions against counting noise where the single-site
baseline ($+1\%$) cannot. The under-35 group is noise-dominated: monthly
rates reach exactly 0\% or 100\% when only one or two patients are
treated, and both models fall below naive, confirming that no
environmental signal can be recovered at this level of aggregation.

Feature importance results from SHAP and LIME are presented in Appendix~B; key findings are summarised below.
Temperature and CO$_2$ emerge as the most consistently important variables across both clinics, while TVOC shows a stronger negative effect in Northern Europe. Site-dependent differences in CO$_2$ direction are discussed in Appendix~B.

\paragraph{Limitations.}
The dataset is limited in size, particularly for the Northern European clinic (3 test months), which restricts statistical power and means reported metrics carry substantial uncertainty. Outcomes are aggregated proportions without patient-level information; factors such as embryo quality, stimulation protocol, and patient history are unobserved and may confound the estimated environmental effects. Similarly, clinic-month-level outcomes may correlate with staffing patterns, lab protocol changes, or seasonal effects unrelated to sensor readings. 
\section{Conclusion and Future Work}
Context-aware features capture patterns that are not visible in simple monthly averages, and this is supported by results on the Asian data and the transfer to Northern Europe. The hierarchical model with partial pooling reaches $R^2=0.86$ for the 35--39 age group, where CO$_2$ and temperature appear consistently important across both clinics. The 40+ group shows only small improvements, and the under-35 group is mostly dominated by noise at this level of aggregation. Overall, the results suggest that routine environmental monitoring is critical to contain useful information about laboratory conditions that goes beyond simple threshold alarms. Future work should focus on larger datasets and adding patient-level and incubator-level data to get more reliable and biologically meaningful results. In addition, counterfactual analysis could be applied in this context to explore “what-if” scenarios, enabling the assessment of how changes in environmental factors (e.g., CO$_2$ levels or temperature) might influence success rates. Such approaches could provide actionable insights for optimizing laboratory conditions and improving clinical outcomes.

\section*{Acknowledgements}
This work was supported by Vinnova, the Swedish Governmental Agency for Innovation Systems [Grant No. 2024-01462]. The funding source had no role in the design, execution, or publication of the study.


\appendix
\onecolumn
\section{Prior Specification and Beta-Binomial Extension}

\paragraph{Choice of prior means $\mu_0$.}
The global mean $\mu_G$ in Eq.~\eqref{eq:hier} is centred on an approximate
prior pregnancy rate $\mu_0$ for each age group. These values are not directly
tabulated in any single source; they are weakly informative approximations
consistent with the ranges reported across European ART registries (clinical
pregnancy rates per fresh autologous cycle). Because $\sigma_G$ is inferred
from data and both Asia and Northern Europe contribute many observations, the posterior
is not sensitive to moderate changes in $\mu_0$.

\begin{table}[h]
\centering
\small
\caption{Weakly informative prior means $\mu_0$ used for $\mu_G$, consistent
with European ART registry ranges~\citep{kupka2016}. These are approximate
starting points; the posterior updates them substantially.}
\begin{tabular}{lccp{5.5cm}}
\toprule
Age group & $\mu_0$ & $\sigma_0$ & Rationale \\
\midrule
$<$35 years    & 0.40 & 0.10 & Approx.\ clinical PR per cycle, European avg \\
35--39 years   & 0.28 & 0.10 & Lower success, wider uncertainty \\
$\geq$40 years & 0.12 & 0.08 & Marked decline; narrow prior, broad posterior \\
\bottomrule
\end{tabular}
\end{table}

\paragraph{Beta-Binomial extension.}
When monthly patient counts $n_{t,c}$ are available, replace Eq.~\eqref{eq:beta}
with $k_{t,c}\sim\mathrm{BetaBinomial}(\mu_{t,c}\kappa,(1-\mu_{t,c})\kappa,n_{t,c})$,
where $k_{t,c}$ is the number of clinical pregnancies and
$\kappa\sim\mathrm{Gamma}(2,0.1)$. This weights each month by how many patients
contributed---the statistically correct treatment for small-count subgroups,
and the most important modelling improvement available for future work.

\section{SHAP and LIME Interpretation}
\label{app:shap}

\begin{figure}[t]
\centering
\includegraphics[width=0.5\textwidth, trim=0cm 20.5cm 14cm 2cm, clip]{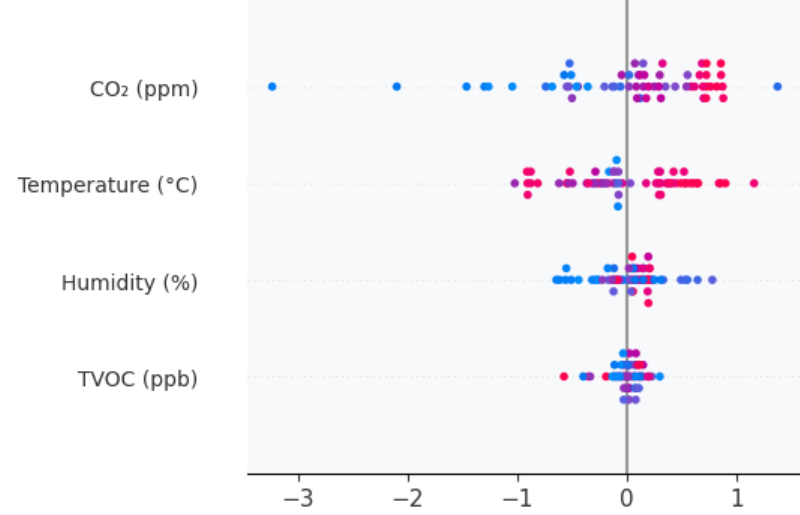}

\includegraphics[width=0.5\textwidth, trim=14cm 20cm 1.8cm 1.8cm, clip]{SHAP.pdf}
\caption{SHAP beeswarm for PR~35--39. \textit{Top}: Asia (61~weeks). \textit{Bottom}:
Northern Europe (14~months). Red = high feature value; blue = low.}
\label{fig:shap}
\end{figure}

\begin{figure}[t]
\centering
\includegraphics[width=0.5\textwidth]{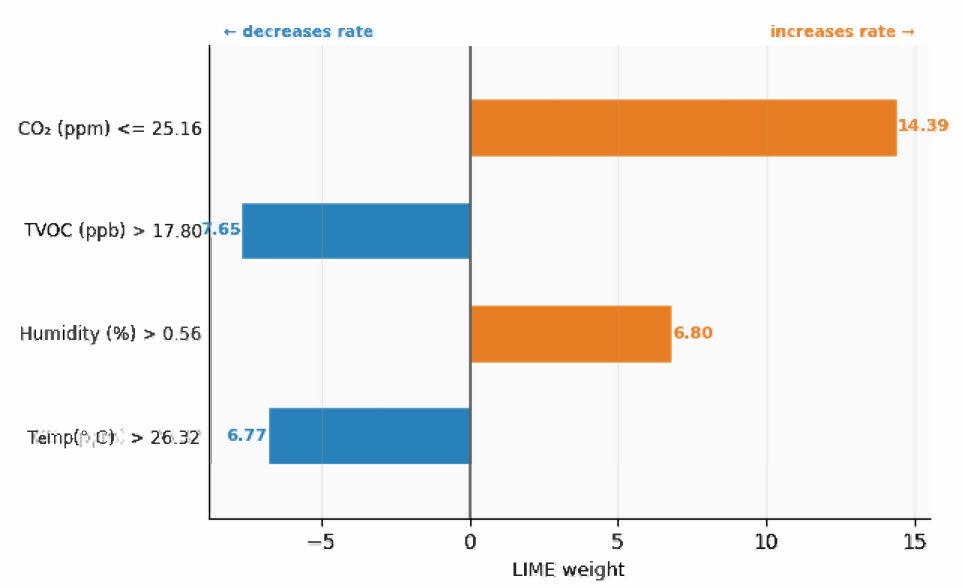}
\caption{LIME explanation for December~2025, PR~35--39 (Northern European clinic).
Orange bars increase the predicted pregnancy rate; blue bars decrease it.}
\label{fig:lime}
\end{figure}

Figures~\ref{fig:shap} and~\ref{fig:lime} provide two complementary
views of how environmental conditions relate to pregnancy rates~\citep{salih2025}.
The SHAP analysis (Figure~\ref{fig:shap}) shows which variables matter
most and in which direction across both clinics.
In the Asian data, higher CO$_2$ values are associated with positive SHAP contributions.
In contrast, in the Northern European clinic, higher TVOC shows a clear negative effect, and
both CO$_2$ and temperature tend to contribute negatively. The opposite directional
effect of CO$_2$ between sites is consistent with the model structure: $\boldsymbol{\beta}$ is shared, but
the posterior is weighted toward the Asian clinic, which provides far more observations.
This site-dependency in CO$_2$ effects likely reflects genuine differences in incubator types,
ventilation systems, or CO$_2$ calibration protocols between the two laboratories, and
warrants investigation with larger datasets. Temperature shows a more consistent pattern
across sites.
LIME (Figure~\ref{fig:lime}) provides a local view by explaining individual predictions.
Averaged over Northern European months, lower CO$_2$ levels consistently contribute
positively, while higher temperature and TVOC reduce predicted pregnancy rates.
Higher humidity shows a positive contribution. LIME highlights how these effects
combine within specific observations, complementing the global view from SHAP.

\end{document}